\documentclass[runningheads]{llncs}

\usepackage[T1]{fontenc}
\def\doi#1{\href{https://doi.org/\detokenize{#1}}{\url{https://doi.org/\detokenize{#1}}}}

\usepackage{graphicx}

\usepackage{hyperref}
\hypersetup{
    colorlinks=true,
    linkcolor=blue,
    filecolor=magenta,
    urlcolor=cyan,
    }
\usepackage[dvipsnames]{xcolor}

\usepackage{listings}
\newcommand{\bftable}{\fontseries{b}\selectfont}

\begin{document}

\title{GaitForeMer: Self-Supervised Pre-Training of Transformers via Human Motion Forecasting for Few-Shot Gait Impairment Severity Estimation}

\titlerunning{GaitForeMer: \underline{Gait} \underline{Fore}casting \& Impairment Estimation Transfor\underline{Mer}}

\author{Mark Endo\inst{1}\orcidID{0000-0003-3218-5807}
\and Kathleen L. Poston\inst{1}\orcidID{0000-0003-3424-7143}
\and Edith V. Sullivan\inst{1}\orcidID{0000-0001-6739-3716}
\and Li Fei-Fei\inst{1}\orcidID{0000-0002-7481-0810}
\and Kilian M. Pohl\inst{1,2}\orcidID{0000-0001-5416-5159}
\and Ehsan Adeli\inst{1}\orcidID{0000-0002-0579-7763}
}
\authorrunning{M. Endo et al.}

\institute{Stanford University, Stanford, CA 94305, USA\\
\email{eadeli@stanford.edu}
\and SRI International, Menlo Park, CA 94025, USA\\
}

\maketitle

\begin{abstract}

Parkinson's disease (PD) is a neurological disorder that has a variety of observable motor-related symptoms such as slow movement, tremor, muscular rigidity, and impaired posture. PD is typically diagnosed by evaluating the severity of motor impairments according to scoring systems such as the Movement Disorder Society Unified Parkinson's Disease Rating Scale (MDS-UPDRS). Automated severity prediction using video recordings of individuals provides a promising route for non-intrusive monitoring of motor impairments. However, the limited size of PD gait data hinders model ability and clinical potential. Because of this clinical data scarcity and inspired by the recent advances in self-supervised large-scale language models like GPT-3, we use human motion forecasting as an effective self-supervised pre-training task for the estimation of motor impairment severity. We introduce \textbf{GaitForeMer}, \underline{Gait} \underline{Fore}casting and impairment estimation transfor\underline{Mer}, which is first pre-trained on public datasets to forecast gait movements and then applied to clinical data to predict MDS-UPDRS gait impairment severity. Our method outperforms previous approaches that rely solely on clinical data by a large margin, achieving an F$_1$ score of 0.76, precision of 0.79, and recall of 0.75. Using GaitForeMer, we show how public human movement data repositories can assist clinical use cases through learning universal motion representations. The code is available at \url{https://github.com/markendo/GaitForeMer}.


\end{abstract}

\section{Introduction}

Large-scale language models \cite{bommasani2021opportunities}, such as the third generation Generative Pre-trained Transformer (GPT-3) \cite{brown2020language} and Contrastive Language-Image Pre-Training (CLIP) \cite{radford2021learning}, have gained great success in solving challenging problems under few or zero-shot settings. They owe their success to self-supervised pre-training on an abundant amount of raw and unlabeled data while completing a downstream task fine-tuned on small datasets. These methods can be of great interest in clinical applications, where data are regularly scarce and limited. For instance, one such application could be automatically estimating motor and gait impairments. This is a crucial step in the early diagnosis of Parkinson's disease (PD).

PD is a chronic, progressive brain disorder with degenerative effects on mobility and muscle control \cite{pd_demaagd}. It is one of the most common neurodegenerative disorders, affecting around 0.57\% of people age 45 and over \cite{pd_prevalence}. Motor symptoms are typically used to diagnose PD, as non-motor symptoms such as cognitive symptoms lack specificity and are complicated to assess \cite{pd_nonmotor}. Most of the prior methods for automated prediction of motor-related PD signs and symptoms use wearable sensors \cite{wearables_Daneault,wearables_hobert,wearables_hssayeni}, but these systems can be expensive and intrusive \cite{lu_first_pd}.

Recent methods have shown that video technology can be a scalable, contactless, and non-intrusive solution for quantifying gait impairment severity \cite{lu_pd_uncertainty}. They use recordings of clinical gait tests from the Movement Disorder Society Unified Parkinson's Disease Rating Scale (MDS-UPDRS) \cite{MDS-UPDRS}, a widely used clinical rating scale for the severity and progression of PD. One of its components is a motor examination where participants walk 10 meters away from and toward an examiner. Specialists assess the severity of motor impairment on a score from 0 to 4. Score 0 indicates no motor impairment, 1 means slight impairment, 2 indicates mild impairment, 3 specifies moderate impairment, and 4 means severe impairment. The goal is automated prediction of this universally-accepted scale.

At the same time, most clinical studies are limited by the difficulty of large-scale data collection, since recruiting patients requires costly multi-site national or international efforts. Recent self-supervised pre-training frameworks have been used to learn useful representations that can be applied to downstream tasks \cite{devlin2019bert,he2020moco,chen2020simclr,he2021maskedautoencoders,ramesh2021zero}. Yet, their translation to clinical applications is under-explored. There is a growing number of 3D human motion capture repositories that can be used for self-supervised pre-training of motion representation learning, the same way GPT-3 \cite{brown2020language} and CLIP \cite{radford2021learning} are trained for language-related tasks. Such pre-trained models can then be adapted for motor impairment estimation.

In this paper, we develop a novel method, {\textbf{GaitForeMer}}, that forecasts motion and gait (pretext task) while estimating impairment severity (downstream task). Our model is based on the recent advances in convolution-free, attention-based Transformer models. GaitForeMer can take advantage of large-scale public datasets on human motion and activity recognition for learning reliable motion representations. To this end, we pre-train the motion representation learning on 3D motion data from the NTU-RGB\texttt{+}D dataset \cite{ntu_rgb+d}. The learned motion representations are then fine-tuned to estimate the MDS-UPDRS gait scores. Our approach outperforms previous methods on MDS-UPDRS score prediction by a large margin. The benefits of GaitForeMer are twofold: (1) we use motion forecasting as a self-supervised pre-training task to learn useful motion features for the task of motor impairment severity prediction; (2) the joint training of motion and MDS-UPDRS score prediction helps improve the model's understanding of motion, therefore leading to enhanced differentiation of motor cues. To the best of our knowledge, we are the first to use motion prediction from natural human activity data as an effective pre-training task for the downstream task of motor impairment severity estimation. We expect this method to be useful in decreasing the reliance on large-scale clinical datasets for the detection of factors defining gait disturbance.

\begin{figure}[t]
\includegraphics[width=\textwidth]{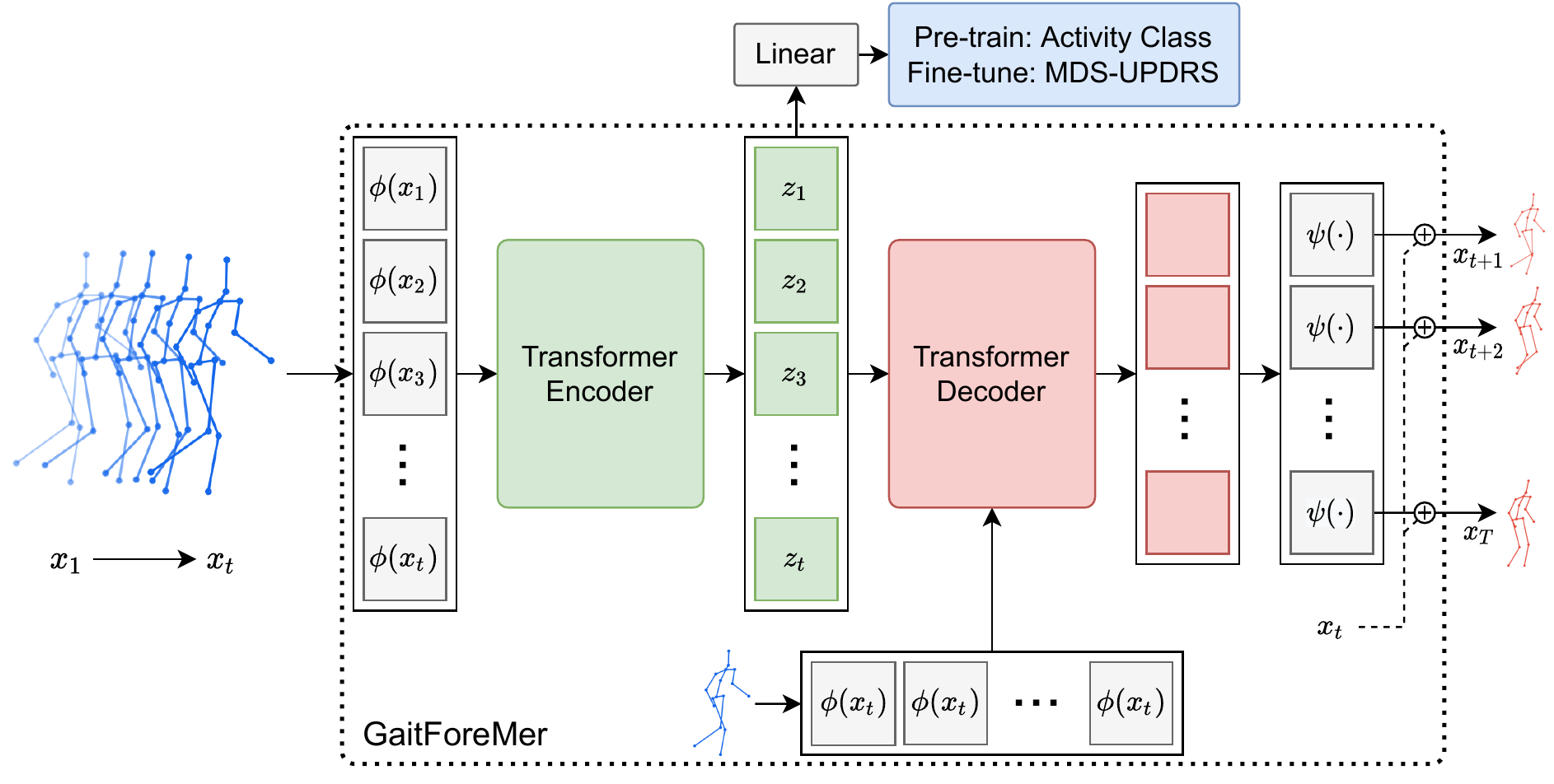}
\caption{The proposed GaitForeMer framework for motor impairment severity estimation. A Transformer model based on body skeletons is first pre-trained on a large dataset jointly for human motion prediction and activity prediction. The model is subsequently fine-tuned on the task of MDS-UPDRS gait score prediction using extracted 3D skeleton sequences from clinical study participants using VIBE (Video Inference for Body pose and shape Estimation) \cite{VIBE}.} \label{fig1}
\end{figure}

\section{GaitForeMer: \underline{Gait} \underline{Fore}casting \& Impairment Estimation Transfor\underline{Mer}}

We first introduce a Transformer model that operates on sequences of 3D human body skeletons for the pre-training task of human motion forecasting, and we subsequently adapt it to the downstream task of MDS-UPDRS gait score estimation. Given a sequence of $t$ 3D skeletons $\mathbf{x}_{1:t}$, we predict the next $M$ skeletons $\mathbf{x}_{t+1:T}$ and the motion class $y$ (either activity or MDS-UPDRS score). We follow the original setup of the pose Transformer model defined in \cite{POTR}. Specifically, our model comprises a skeleton encoding neural network $\phi$, a Transformer encoder and decoder, a pose decoding neural network $\psi$ that reconstructs the 3D poses, and a linear classifier for MDS-UPDRS score/activity prediction. The model takes $\mathbf{x}_{1:t}$ as input, and the skeleton encoding network $\phi$ embeds this joint data into dimension $D$ for each skeleton vector. Then, the Transformer encoder takes in this sequence of skeleton embeddings aggregated with positional embeddings and computes a latent representation $\mathbf{z}_{1:t}$ using $L$ multi-head self-attention layers. The outputs of the encoder embeddings, $\mathbf{z}_{1:t}$, are fed into a single linear layer to predict class probabilities. The classification loss $L_{c}$ (a multi-class cross-entropy) is used to train activity or score prediction. Note that the class prediction uses the motion representation (i.e., the latent space of the Transformer) as input and outputs the activity classes in the pre-training stage and MDS-UPDRS scores in the downstream task.

In addition, the encoder outputs $\mathbf{z}_{1:t}$ and a query sequence $\mathbf{q}_{1:M}$ are fed into the Transformer decoder. The query sequence is filled with the last element of the input sequence $\mathbf{x}_t$. The decoder uses $L$ multi-head self- and encoder-decoder attention layers. The output of the decoder is fed into a skeleton decoding network $\psi$ to generate the future skeleton predictions $\mathbf{y}_{t+1:T}$. The motion forecasting branch is trained using a layerwise loss calculated as:

$$L_l = \frac{1}{M \cdot N} \sum_{m=t+1}^{T} \|\hat{\mathbf{x}}^l_m - \mathbf{x}^*_m\|_1,$$
where $\hat{\mathbf{x}}^l_m$ is the predicted sequence of $N$-dimensional skeleton vectors at layer $l$ of the Transformer decoder, and $\mathbf{x}^*_m$ is the ground-truth future skeleton. The motion forecasting loss $L_{f}$ is then computed by averaging the layerwise loss over all decoder layers $L_l$. See Figure \ref{fig1} for an overview of the model architecture.

\subsection{Pre-training Procedure}

For pre-training, we jointly train the activity and motion branches of the GaitForeMer. For the classification loss $L_{c}$, we use a standard categorical cross-entropy loss. The final loss is calculated as $L_{pre} = L_{c} + L_{f}$, where there is an equal weighting of the two different losses. We train the model for 100 epochs using an initial learning rate of 0.0001.

\subsection{Fine-tuning Procedure}

For our downstream task of MDS-UPDRS score prediction, we initialize the model using the learned weights from pre-training. We set the classification loss $L_{c}$ as a weighted categorical cross-entropy loss since there is a significant class imbalance in the clinical data. We experiment with a variety of training procedures for fine-tuning the model. In one setup, we solely fine-tune the class prediction branch by setting $L_{fine} = L_{c}$. In another setup, we fine-tune both the class prediction branch and the motion prediction branch by setting $L_{fine} = L_{c} + L_{f}$. We also experiment with first fine-tuning both branches for 50 epochs then additionally solely fine-tuning the class prediction branch for 50 epochs. All fine-tuning setups are trained for 100 epochs using an initial learning rate of 0.0001.

\subsection{Baselines}

We compare our GaitForeMer method to a similar setup without pre-training on the NTU RGB\texttt{+}D dataset as well as various other motion impairment severity estimation models for MDS-UPDRS score prediction. The GaitForeMer model trained from scratch (GaitForeMer-Scratch) uses the loss function $L = L_{c} + L_{f}$ and follows the same configuration as the fine-tuning setups except it is trained for an additional 100 epochs. Hybrid Ordinal Focal DDNet (OF-DDNet) \cite{lu_pd_uncertainty} uses a Double-Features Double-Motion Network with a hybrid ordinal-focal objective. This previous method has shown promising results on this MDS-UPDRS dataset. Spatial-Temporal Graph Convolutional Network (ST-GCN) is a graphical approach for learning spatial and temporal characteristics that can be used for action recognition \cite{st-gcn}. For this method, we add slow and fast motion features and pass the input through a Graph Attention Network (GAT) \cite{gat} layer first to allow for additional spatial message passing. DeepRank \cite{deeprank} is a ranking CNN, and the Support Vector Machine (SVM) \cite{svm} is using the raw 3D joints.

\section{Datasets}

In this work, we use a clinical dataset for estimation of gait impairment severity from MDS-UPDRS videos and a public 3D human gait dataset for pre-training the motion forecasting component. Both datasets are described below. 

\subsection{NTU RGB\texttt{+}D Dataset}
We use NTU RGB\texttt{+}D \cite{ntu_rgb+d} to pre-train our GaitForeMer model. This dataset includes 56,880 video samples with 60 action classes. We use the skeletal data containing 3D coordinates of 25 joints and class labels. We pre-train our model using the joints as input and the activity labels for supervision of the activity branch. The activity branch is the same as our MDS-UPDRS branch (see Fig. \ref{fig1}), except that during pre-training we train the linear layers to predict the activity class in the NTU RGB\texttt{+}D dataset.

\subsection{MDS-UPDRS Dataset}

For the downstream task of gait impairment severity estimation, we use the MDS-UPDRS dataset defined in \cite{lu_pd_uncertainty}. This dataset contains video recordings of MDS-UPDRS exams from 54 participants. Following previously published protocols \cite{Poston-medicationstate}, all participants are recorded during the off-medication state.
During the examinations, participants are recorded walking towards and away from the camera twice at 30 frames per second.
Each sample is scored by three board-certified movement disorders neurologists and we use the majority vote among the raters as the ground-truth score, randomly breaking ties. Note that, in this work, we do not aim to model the uncertainty among raters. The raters score the videos on a scale from 0 to 4 based on MDS-UPDRS Section 3.10 \cite{goetz2008mds}. In our work, we combine scores 3 and 4 due to the difficulty of obtaining video recordings for participants with extreme motor impairment. The data setup and protocols are the same as in \cite{lu_pd_uncertainty}, except in their previous work two sets of scores were counted from one of the neurologists. Using the gait recordings as input, we use Video Inference for Body Pose and Shape Estimation (VIBE) to extract 3D skeletons \cite{VIBE}. This joint data is then preprocessed by normalization and splitting of samples into clips of 100 frames each. We then use these clips for estimating motor impairment severity.

\section{Experiments}

In this section, we first evaluate how motion forecasting helps improve a system estimating the MDS-UPDRS scores. We compare our results with several baselines (Section \ref{Section 4.1}). We then evaluate how the fine-tuning strategy contributes to better results (Section \ref{Section 4.2}). We further experiment on how our few-shot learning paradigm can be adopted for clinical approaches using pre-training (Section \ref{Section 4.3}). Qualitative results on motion forecasting of PD patients validate that GaitForeMer is able to learn good motion representations (Section \ref{Section 4.4}).

\begin{table}[t]
    \caption{Comparison with baseline methods. Performance is evaluated using macro F$_1$ score, precision, and recall. We find that pre-training results in significantly improved performance over training from scratch and the other methods. $\natural$ refers to results directly cited from \cite{lu_pd_uncertainty}. * indicates statistical difference at $(p<0.05)$ compared with our method, measured by the Wilcoxon signed rank test \cite{wilcoxon1992individual}. Note that this is a 4-class classification problem and hence 0.25 recall implies a random classifier. Best results are in bold. See text for details about compared methods.}
    \label{tab:baselines-comparion}
    \setlength{\tabcolsep}{8pt}
    \centering
    \begin{tabular}{lccccc}
    \hline
    Method & F$_1$ & Pre & Rec \\
    \hline
    GaitForeMer (Ours) & \bftable 0.76 & \bftable 0.79 & \bftable 0.75 \\
    GaitForeMer-Scratch (Ours) & 0.60 & 0.64 & 0.58 \\
    OF-DDNet$^\natural$* \cite{lu_pd_uncertainty} & 0.58 & 0.59 & 0.58\\
    ST-GCN \cite{st-gcn}* & 0.52 & 0.55 & 0.52\\
    DeepRank$^\natural$* \cite{deeprank} & 0.56 & 0.53 & 0.58 \\
    SVM$^\natural$* \cite{svm} & 0.44 & 0.49 & 0.40 \\
    \hline
    \end{tabular}
\end{table}

\subsection{Using Motion Forecasting as an Effective Pre-training Task} \label{Section 4.1}

We investigate the efficacy of using human motion forecasting as a self-supervised pre-training task for the downstream task of motor impairment severity estimation. We evaluate each model using macro F$_1$ score, precision, and recall. These metrics are calculated on a per subject level with leave-one-out-cross-validation settings. We compare our GaitForeMer method to baseline methods in Table \ref{tab:baselines-comparion}.

We find that our GaitForeMer method pre-trained on the NTU RGB\texttt{+}D dataset results in improved performance over training the model from scratch and all baselines trained on the MDS-UPDRS dataset. Our best setup achieves an F$_1$ score of 0.76, precision of 0.79, and recall of 0.75. In comparison, training the model from scratch results in an F$_1$ score of 0.60, precision of 0.64, and recall of 0.58, which is still superior to other baselines. The OF-DDNet baseline (previous state-of-the-art approach in MDS-UPDRS score prediction) has an F$_1$ score of 0.58, precision of 0.59, and recall of 0.58.

\begin{table}[t]
    \caption{Comparison of different training/fine-tuning strategies of our method (ablation study on fine-tuning strategy). Performance is evaluated using macro F$_1$ score, precision, and recall. We find that first fine-tuning both branches (forecasting and score prediction) then additionally fine-tuning the score prediction branch yields best results.}
    \label{tab:fineune-setups}
    \setlength{\tabcolsep}{8pt}
    \centering
    \begin{tabular}{llccc}
    \hline
    Pre-trained & Fine-tune strategy & F$_1$ & Pre & Rec \\
    \hline
    Yes & Both branches then class branch & \bftable 0.76 & \bftable 0.79 & \bftable 0.75 \\
    Yes & Both branches & 0.72 & 0.75 & 0.71 \\
    Yes & Class branch & 0.66 & 0.72 & 0.63 \\
    No & & 0.60 & 0.64 & 0.58 \\
    \hline
    \end{tabular}
\end{table}

\subsection{Evaluating Fine-tuning Strategies} \label{Section 4.2}

We experiment with various fine-tuning strategies in order to evaluate different approaches for training our GaitForeMer method. Our best approach of first fine-tuning both the class prediction and motion prediction branches then solely fine-tuning the class prediction branch achieves an F$_1$ score of 0.76, precision of 0.79, and recall of 0.75. Another approach of fine-tuning both branches achieves an F$_1$ score of 0.72, precision of 0.75, and recall of 0.71. We observe that solely fine-tuning the class branch results in worse performance than also training the motion branch with an F$_1$ score of 0.66, precision of 0.72, and recall of 0.63. The relatively poor performance could be due to the data shift between the NTU RGB\texttt{+}D and MDS-UPDRS datasets that requires training of the motion forecasting branch. Results are shown in Table \ref{tab:fineune-setups}.

\begin{figure}[b]
\centering
\includegraphics[width=.65\textwidth]{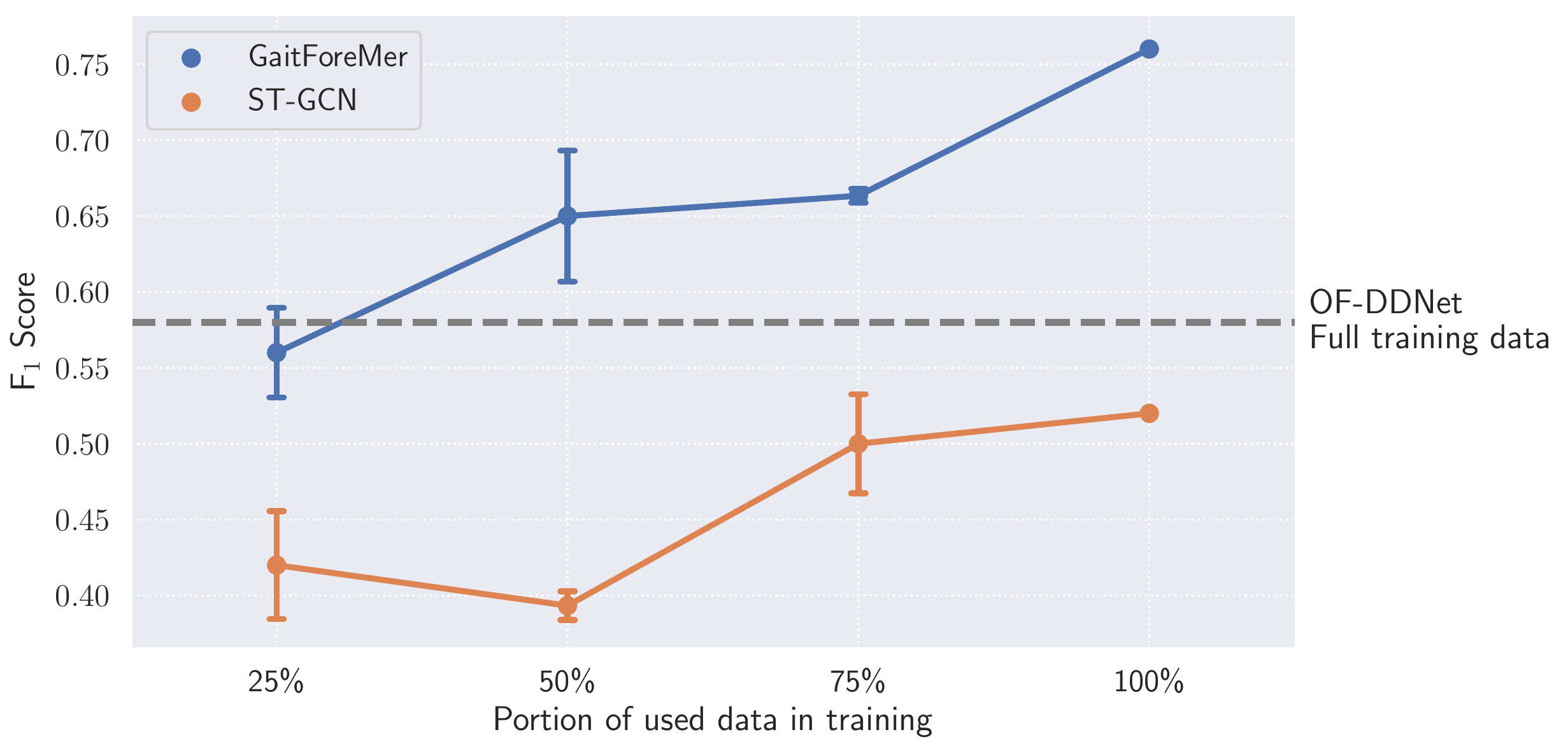}
\caption{Few-shot performance of GaitForeMer compared to ST-GCN with different portions of data used in training. Error bars represent standard deviation across 3 runs. Our GaitForeMer method using only 25\% of training data maintains comparable performance to the second-best performing method with full training data (OF-DDNet).} \label{few_shot_figure}
\end{figure}

\subsection{Few-Shot Estimation of Gait Scores} \label{Section 4.3}

To better understand the few-shot capabilities of GaitForeMer, we experiment with limiting the training dataset size and evaluating performance compared to ST-GCN. We sample either 25\%, 50\%, or 75\% of the data (analogous to 13, 26, or 39 videos) for training in each fold, preserving the same samples across the two methods. We maintain class balance by sampling one-fourth of the required subsamples from each class when permitted. We resample and run each method three times. The results are illustrated in Figure \ref{few_shot_figure}.

We find that our GaitForeMer method maintains relatively strong performance with only a fraction of the data. 
GaitForeMer with access to 25\% of training data achieves an average F$_1$ score of 0.56, which is higher than ST-GCN using 100\% of training data with an F$_1$ score of 0.52 and comparable to OF-DDNet (second-best performing method) using 100\% training data with an F$_1$ score of 0.58. This shows the power of using motion forecasting as a self-supervised pre-training task for few-shot gait impairment severity estimation.

\subsection{Motion Forecasting Visualization} \label{Section 4.4}

\begin{figure}[t]
\includegraphics[width=\textwidth]{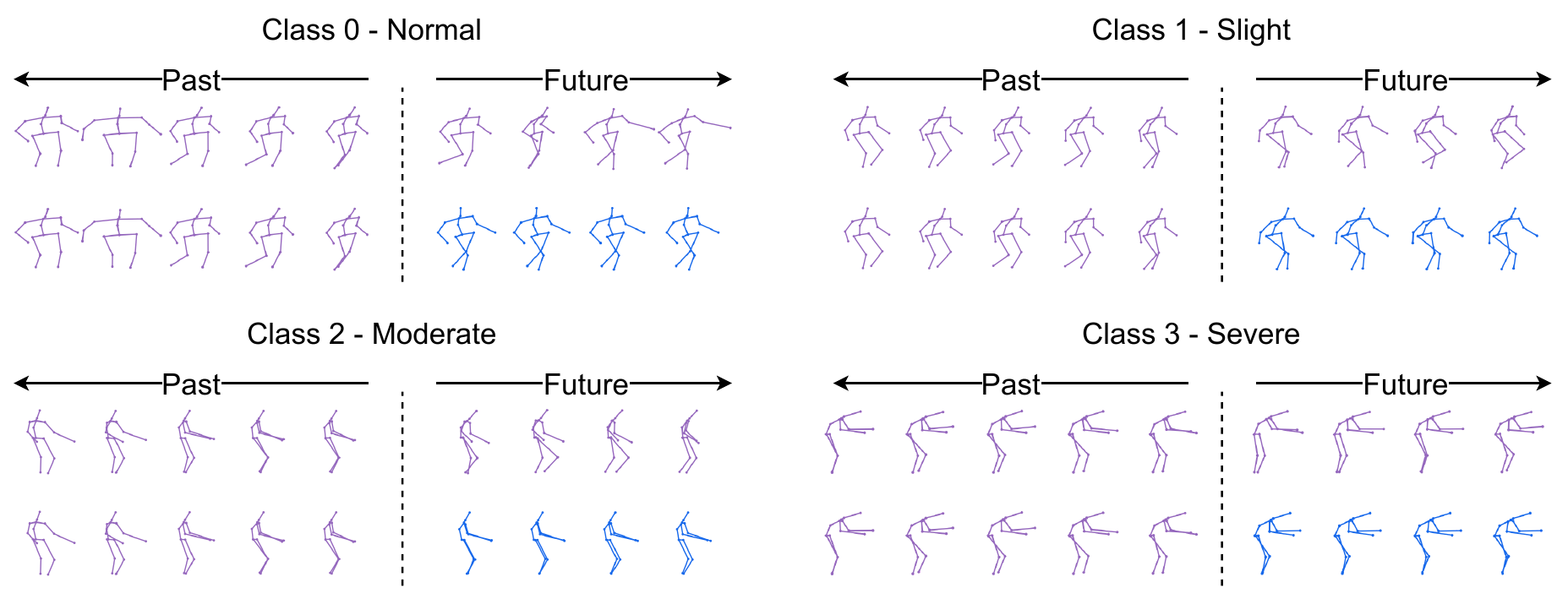}
\caption{Visualization of human motion forecasting for different levels of motor impairment severity. The \textcolor{Plum}{purple} skeletons are {ground-truth} data, and the \textcolor{blue}{blue} ones are predictions from GaitForeMer with fine-tuning both branches.} \label{visualization_figure}
\end{figure}

In Figure \ref{visualization_figure}, we visualize the predicted outputs of the GaitForeMer model jointly trained on MDS-UPDRS score prediction and motion forecasting. Although accurate pose forecasting is not necessary for the prediction of MDS-UPDRS scores, it can help demonstrate the utility of learned motion features. Qualitatively, we see that the predicted poses most closely match the ground-truth at the beginning of the output. This might be because using the last input entry $\mathbf{x}_t$ as the query sequence $\mathbf{q}_{1:M}$ helps the prediction in the short term \cite{POTR}. A larger error exists for longer horizons where the outputted poses become less similar to the query sequence. In addition, the non-autoregressive approach of GaitForeMer can lead to an increased accumulation of error.

Clinically, the results illustrated in Figure \ref{visualization_figure} show normal movement behavior in class 0 (normal), while classes 1 and 2 show increased stiffness, decreased mobility, and reduced arm swing and pedal motion. Participants in class 3 are imbalanced and require assistive devices for safe walking. These results verify that the forecasting module is able to properly predict future motion that encodes motor impairments.

\section{Conclusion}
Herein, we presented a model, GaitForeMer, based on transformers for forecasting human motion from video data that we use to predict MDS-UPDRS gait impairment severity scores. We found that human motion forecasting serves as an effective pre-training task to learn useful motion features that can subsequently be applied to the task of motor impairment severity estimation, even in few-shot settings. The pre-trained GaitForeMer outperformed training from scratch and other methods for motor impairment severity estimation that solely use the MDS-UPDRS dataset for training. Our approach demonstrates the utility of using motion pre-training tasks in data-limited settings.

\subsubsection{Acknowledgements} 
This work was supported in part by NIH grants (AA010723, NS115114, P30AG066515), the Michael J Fox Foundation for Parkinson’s Research, UST (a Stanford AI Lab alliance member), and the Stanford Institute for Human-Centered Artificial Intelligence (HAI) Google Cloud credits.

%
%
%
\bibliographystyle{splncs04}
\bibliography{paper398.bib}

\end{document}